%% file: iclr2025_conference.tex
\documentclass{article} 
\usepackage{tccml_iclr2025_conference,times}
\input{math_commands.tex}

\usepackage{hyperref}
\usepackage{url}
\usepackage{subcaption}
\usepackage[pdftex]{graphicx}

\usepackage{enumitem}

\usepackage[latin1]{inputenc}
\usepackage{amssymb,amsmath,array}
\usepackage{booktabs}       
\usepackage{amsfonts}       
\usepackage{nicefrac}       
\usepackage{microtype}      
\usepackage{comment}
\usepackage{todonotes}
\usepackage{subcaption}

\title{Prototype-enhanced prediction in graph \\ neural networks for climate applications}

\author{Nawid Keshtmand, Elena Fillola, Jeffrey Nicholas Clark, \\
\textbf{Raul Santos-Rodriguez} \& \textbf{Matthew Rigby} \\
University of Bristol\\
Bristol, England \\
\texttt{yl18410@bristol.ac.uk}
}

\begin{document}

\maketitle

\begin{abstract}
Data-driven emulators are increasingly being used to learn and emulate physics-based simulations, reducing computational expense and run time. Here, we present a structured way to improve the quality of these high-dimensional emulated outputs, through the use of prototypes: an approximation of the emulator's output passed as an input, which informs the model and leads to better predictions. We demonstrate our approach to emulate atmospheric dispersion, key for greenhouse gas emissions monitoring, by comparing a baseline model to models trained using prototypes as an additional input. The prototype models achieve better performance, even with few prototypes and even if they are chosen at random, but we show that choosing the prototypes through data-driven methods (k-means) can lead to almost 10\% increased performance in some metrics.
  
\end{abstract}

\section{Introduction}
Earth system modeling often relies on complex physics-based simulations, which, are computationally expensive and can limit their applicability for real-time decision-making or large-scale datasets. Machine learning (ML) models are increasingly being used to accelerate, or fully replace, these expensive simulations \citep{fillola2023machine, vinuesa2022enhancing,bhowmik2022deepclouds}. However, ML-driven emulators are often required to predict a high-dimensional output state from a high-dimensional input state, and exploring techniques to improve their performance is key to ensuring they can serve as reliable surrogates for physics-based simulations.

Here, we introduce the concept of `prototypes': an approximation of the emulator's output provided as an input, serving as a prior or guide to help the emulator produce higher-quality predictions.  

We demonstrate the utility of prototypes in monitoring greenhouse gas (GHG) emissions. Traditionally, countries report their emissions through national greenhouse gas inventories (NGHGIs), compiled through a bottom-up approach based on socioeconomic activity data and emissions factors \citep{olaguer2016atmospheric}.
An alternative is the top-down inverse modeling approach, which uses atmospheric observations to estimate GHG emissions. This method employs atmospheric dispersion models to solve for atmospheric transport, with Lagrangian Particle Dispersion Models (LPDMs) being among the most popular for this application \cite{jacob2022quantifying}. When used in inverse modelling systems, LPDMs simulate gas `particles' moving backward in time from a GHG measurement location, generating a 2D histogram - or `footprint'- that identifies the contribution of upwind fluxes to the observed GHG concentrations \citep{hegarty2013evaluation} (see examples in Fig. \ref{fig:visual_prototypes}). With one LPDM run required for each measurement and taking $\sim$20 CPU-minutes, these methods face challenges with the vast increase in GHG data from new satellite-based instruments, now totalling thousands of measurements every day. The computational cost of generating footprints creates a bottleneck in emission estimation. Here, we aim to improve an existing LPDM-emulator by \cite{fillola2023accelerating}, which uses an architecture based on Graph Neural Networks (GNNs)\citep{battaglia2018relational,bacciu2020gentle} and can predict satellite GHG measurement footprints based on meteorological and topographical features $\sim$1000$\times$ faster than the physics-based dispersion model.


Our contributions include: 1) the development of a prototype-based approach for GNN model predictions and, 2)
experimental validation of this method in the context of GHG emissions footprint prediction.
We show that incorporating prototypes improves the model's performance across several metrics compared to the same model without prototypes.

\section{Method}
\textbf{Architecture}
We use the same Encoder-Processor-Decoder architecture as \cite{fillola2023accelerating}, which is similar to the approach used in meteorological applications \citep{keisler2022forecasting,lam2022graphcast,pfaff2020learning}. The inputs, including meteorological and topographical data, are arranged in a latitude-longitude grid that represents the native data space. In the Encoder this data gets encoded into a triangular, lower-resolution mesh, and in the Processor each mesh node is updated iteratively with a GNN block \citep{battaglia2018relational}, using information from neighbouring nodes. The Decoder then maps the mesh to the original data space, emulating the footprint value at each grid node. For a visual representation of the architecture and more details, see \cite{fillola2023accelerating}.

\textbf{Data}
We use footprints produced by NAME, the UK's Met Office LPDM \citep{jones2007uk} for GOSAT measurements (Greenhouse gases Observing SATellite \citep{parker2020decade}) over Brazil for 2014-16, used originally by \cite{tunnicliffe2020quantifying}. The footprints have a resolution of 0.352$^\circ$$\times$0.234$^\circ$ ($\sim$25$\times$35km in mid-latitudes) and have been cropped to a size of 50x50 gridcells ($\sim$1000$\times$1500km), centered around the coordinates of the satellite measurement. They aggregate gas dispersion over 30 days prior to each observation. The inputs are meteorological features at a range of heights, plus information about the topography and coordinates (see \cite{fillola2023accelerating}). We train on data for 2014-2015 ($\sim$11000 data points after sampling by using every third sample in the dataset as samples close in time would be similar to one another), and test on data for 2016 ($\sim$7100 data points after sampling). 

Here, we compare the performance of the original model \citep{fillola2023accelerating}, with the same model with an additional input feature: a prototype.

\begin{figure}[h!]
\centering

     \begin{subfigure}[h]{1.0\textwidth}
         \includegraphics[width=\textwidth]{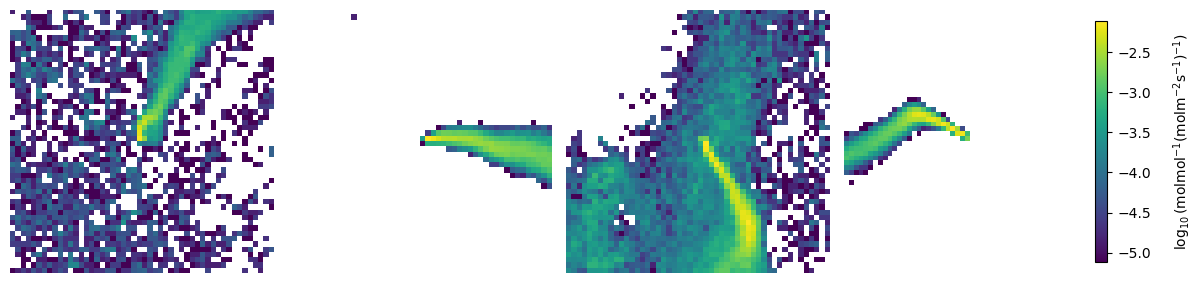}
         \caption{Prototype set chosen by a human expert}
         \label{fig:human_prototypes}
     \end{subfigure}
     \hfill    
     \begin{subfigure}[h]{1.0\textwidth}
         \centering
         \includegraphics[width=\textwidth]{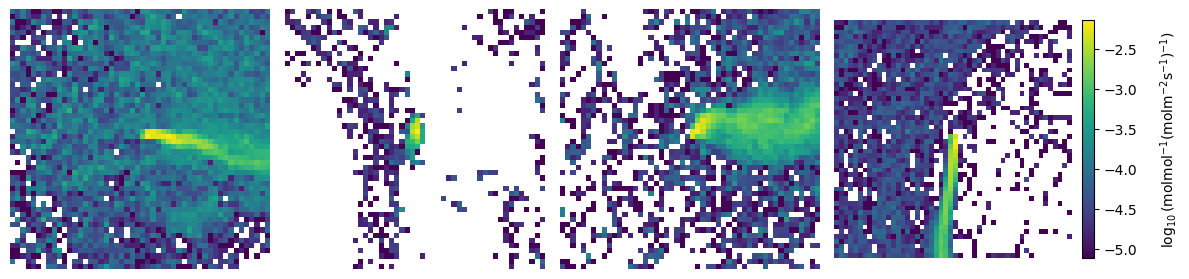}      
         \caption{Prototype set chosen by k-means}
         \label{fig:k_mean_prototypes}
     \end{subfigure}
     \caption{Examples of two prototype sets, with size n = 4. Prototypes are chosen from the true footprints in the training dataset. During training and testing, the footprint from the set closest to the true label is passed as an input to improve prediction. A human expert (a) selects footprints that have distinct wind directions, whereas k-means (b) chooses footprints based on the data distribution.}\label{fig:visual_prototypes}
\end{figure}

\textbf{Prototype Design}
The prototype, i.e. an initial inexpensive approximation of the output, can be produced in several ways: low-cost calculations based on the inputs, using artificial samples, or using selected examples from the training data.  
\cite{he2023footnet} designed a CNN-based emulator for an LPDM, but they operate at a smaller domain and higher resolution (400$\times$400km domain, km-scale resolution). At this scale, Gaussian plumes can be used as prototypes: a simple mathematical approximation of dispersion generated using mainly wind direction and speed \citep{he2023footnet}. They demonstrate how the addition of this input improves the performance of their model. However, Gaussian plumes do not generalize to large domains (e.g. continental scales) and therefore we choose an alternative approach: we select $n$ footprints from the training data to be prototypes. Each training sample then gets assigned one of these prototypes as an additional input, to aid prediction. There are two main steps:
\begin{enumerate}[noitemsep,topsep=0pt,leftmargin=*]
\item Choosing the set of prototypes: the chosen samples should be representative of the range of potential outputs. Here, we demonstrate two methods to curate the prototype set: 
a human expert, and a data-driven approach. An atmospheric dispersion expert chooses manually $n$ footprints, aiming to cover a wide range of different conditions, such as where the upwind areas of the footprint are one of the four main cardinal directions (Fig. \ref{fig:human_prototypes}).  For comparison, we also train a model where $n=4$ prototype footprints are chosen at random. For the data-driven approach, we implement k-means clustering with $n$ clusters, and use the footprint closest to the center of each group as prototype (Fig. \ref{fig:k_mean_prototypes}). If there is a bias in the data distribution (for example, in our dataset for South America many footprints have a East direction due to prevailing winds), the data-driven prototypes will reflect it (e.g. see Fig. \ref{fig:k_mean_prototypes}).
\item Assigning each sample the best-fitting prototype from the prototype set: to demonstrate the impact of prototypes alone, we show an `oracle case' where we have access to the true footprints to assign prototypes. Each footprint in the dataset gets assigned the prototype with the lowest $L_{2}$ distance in a lower dimensional (64) PCA space. In deployment, the true footprint to be predicted is unknown, so the prototype for a particular data point would need to be chosen using the inputs (e.g. using a classifier). 

\end{enumerate}


\textbf{Training details} All models are trained for 100 epochs with a batch size of 5, using the Adam optimizer with a learning rate of $5e^{-5}$. We train and run the model on a 32GB NVIDIA V100 GPU which takes approximately 4 hours. Each model is trained for three different random seeds.

\textbf{Performance metrics}
Test set footprints were evaluated quantitatively and qualitatively. Quantitatively, we calculated Mean Squared Error (MSE) on a pixel-by-pixel basis, and intersection over union (IoU), to measure spatial overlap between the true and the predicted footprints: $\text{IoU} = \frac{\text{Area of Overlap between predicted and true images}}{\text{Area of Union of the predicted and true images}} $ \citep{rahman2016optimizing}.

    \label{eqn:iou}

\begin{figure}[h!]
\centering
     \begin{subfigure}[h]{0.82\textwidth}
         \caption{Intersection over Union (IoU) score for different prototype sets (Higher is better)}
         \includegraphics[width=\textwidth]{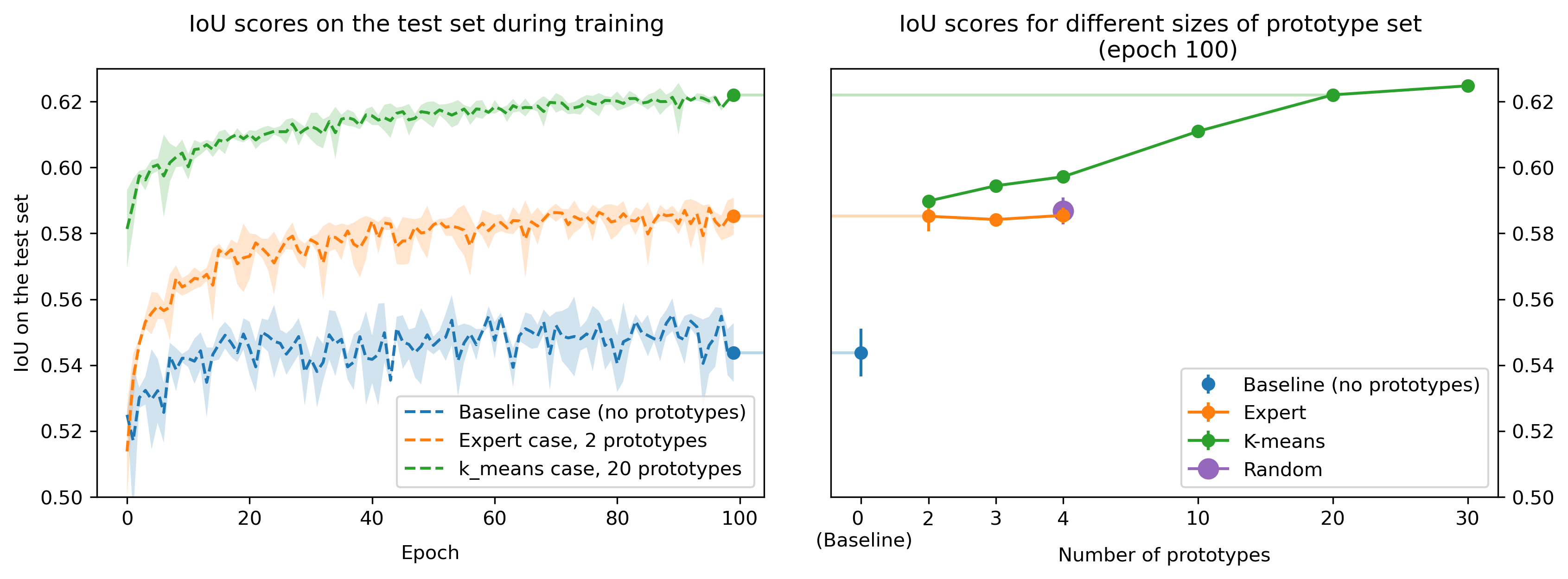}
         \label{fig:IOU_condensed}
         \vspace{-1.07\baselineskip}
     \end{subfigure}
     
     \begin{subfigure}[h]{0.85\textwidth}
         \centering
         \caption{Mean Squared Error (MSE) score for different prototype sets (Lower is better)}
         \includegraphics[width=\textwidth]{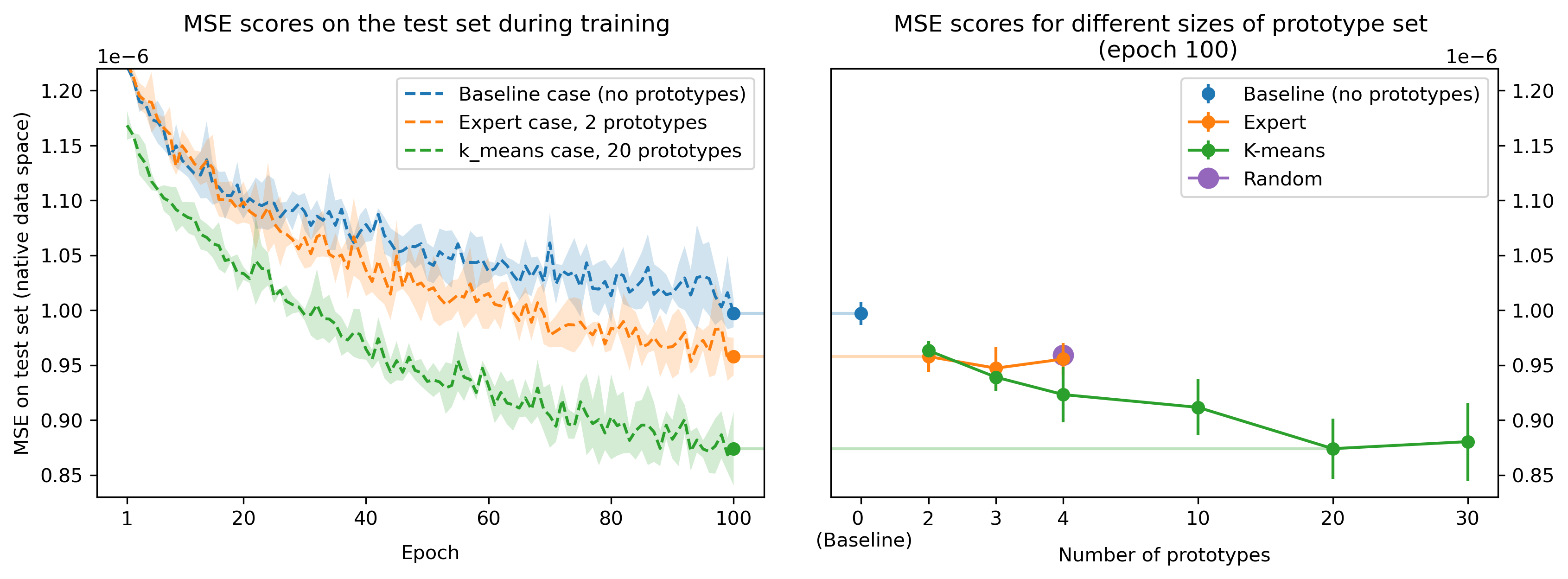}    
         
         \label{fig:MSE_condensed}
     \end{subfigure}
     \caption{Comparison of different models, showing metrics (a) Intersection over Union and (b) Mean Squared Error. The left panels show training curves for three selected models averaged over three seeds (shaded area shows standard distribution). The right panels show the mean score at epoch 100 for different prototype sets and sizes. Error bars show standard deviation across three seeds. }
     \label{fig:results}
\end{figure}
\vspace*{-0.2cm}

\section{Results and discussion}
\textbf{Quantitative analysis}
We compare the performance of a baseline model (0 prototypes, same as \cite{fillola2023accelerating}) with our prototype approach for a range of $n$ prototypes, using a) prototype sets chosen by an expert and b) prototype sets chosen through k-means (Fig. \ref{fig:results}). Using at least two prototypes, through any method, improves the performance on the baseline.

We find that the data-driven approach, utilising prototypes selected with k-means, achieves better improvements on the baseline than the expert-driven method. This improvement becomes more significant with a higher number of prototypes, 
with the k-means model with $n=20$ prototypes producing a mean IoU score 8\% higher than the baseline, and 4\% higher than $n=2$ expert prototypes. We chose to compare the $n=2$ expert prototypes and $n=20$ k-means prototypes as increasing the number of expert prototypes further or increasing the number of k-means prototypes further does not lead to a   significant enhancement in the performance as shown by the small difference between $n=20$ and $n=30$, indicating that $n\simeq20$ prototypes are able to sufficiently represent the data space.  Therefore $n=2$ expert prototypes and $n=20$ k-means prototypes would be a good option for a practitioner aiming to maximise performance whilst minimizing computational cost.

The random experiment, where $n=4$ footprints are chosen arbitrarily to be the prototype set, achieves comparable performance to the expert-chosen set, showing that manually curating the set might not provide the expected advantage.

\begin{figure}[h]
    \centering\includegraphics[width=1.0\columnwidth]{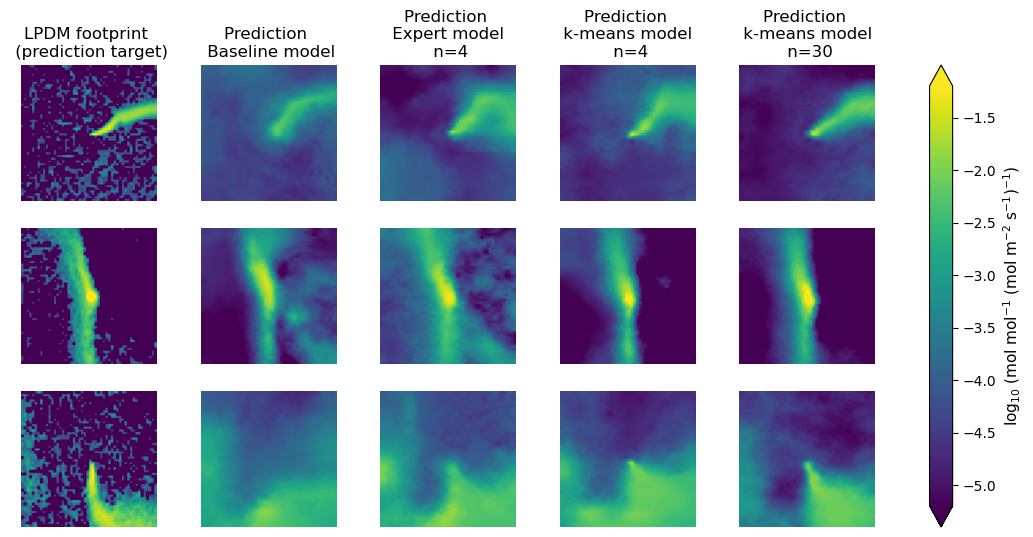}
    \caption{For three random samples from the test set (each row), comparison of the true footprint (first column) with the baseline output not utilising prototypes (second column) and three prototype models. Models are shown ordered by increasing performance, from left to right. 
    }
    \label{fig:qualititative_predictions}
\end{figure}

\textbf{Qualitative analysis}
The predictions made by the baseline model without prototypes (Fig \ref{fig:qualititative_predictions}, second column) capture the main upwind direction of the true footprint (Fig \ref{fig:qualititative_predictions}, first column). However, the predictions tend to be smoothed compared to the ``true'' footprints, and do not always represent well the higher values. Predictions by models that use prototypes produce sharper footprints that better capture the wind direction and high values, particularly in complex meteorological scenarios.

\newpage
\section{Conclusion and Future work}

Here, we propose using prototypes as an additional input to emulators of physical processes. We demonstrate a structured, easy-to-implement approach to improve GNN-based footprint predictions: footprints predicted using prototypes achieve better IoU and MSE scores, capture the upwind direction and shape better, and are less prone to over-smoothing (Fig \ref{fig:qualititative_predictions}). This work enhances efficient GHG atmospheric transport simulators with machine learning, making them practical for real-world use and enabling analysis of GHG sources and sinks with vastly more data.
We demonstrate an oracle set-up, where we assign the optimal prototypes. In practical applications, prototype assignments would need to utilize input features, rather than the target footprint itself. This could be achieved by designing a classifier that chooses a prototype for each sample based on meteorological features.  

Future work will focus on developing a deployment pipeline, developing prototype classifiers and exploring how they impact the performance. It is also key to explore other applications of prototypes, collaborating with domain experts and those developing emulators, since it is likely that other physical emulators could benefit from incorporating prototypes into their emulation pipeline. Increasing the performance and reliability of these emulators will in turn lead to wider and faster adoption of systems that are key for climate modeling, such as near-real time greenhouse gas emissions monitoring.

\subsubsection*{Acknowledgments}
NK, JNC, and MR are funded by the Self-Learning Digital Twins for Sustainable Land Management grant [grant number EP/Y00597X/1].
EFM is funded by a
Google PhD Fellowship. RSR is
funded by the UKRI Turing AI Fellowship [grant
number EP/V024817/1].

\bibliography{iclr2025_conference}
\bibliographystyle{iclr2025_conference}


\end{document}

%% file: math_commands.tex
\usepackage{amsmath,amsfonts,bm}

\def\eqref#1{equation~\ref{#1}}

\def\1{\bm{1}}

\DeclareMathAlphabet{\mathsfit}{\encodingdefault}{\sfdefault}{m}{sl}
\SetMathAlphabet{\mathsfit}{bold}{\encodingdefault}{\sfdefault}{bx}{n}

